# Improved Total Variation based Image Compressive Sensing Recovery by Nonlocal Regularization

*Jian Zhang, Shaohui Liu, Debin Zhao*
School of Computer Science and Technology
Harbin Institute of Technology, Harbin, China
{jzhangcs, shliu, dbzhao}@hit.edu.cn

*Ruiqin Xiong, Siwei Ma*
Institute of Digital Media,
Peking University, Beijing, China
{rqxiong, swma}@pku.edu.cn

*Abstract*—Recently, total variation (TV) based minimization algorithms have achieved great success in compressive sensing (CS) recovery for natural images due to its virtue of preserving edges. However, the use of TV is not able to recover the fine details and textures, and often suffers from undesirable staircase artifact. To reduce these effects, this paper presents an improved TV based image CS recovery algorithm by introducing a new nonlocal regularization constraint into CS optimization problem. The nonlocal regularization is built on the well known nonlocal means (NLM) filtering and takes advantage of self-similarity in images, which helps to suppress the staircase effect and restore the fine details. Furthermore, an efficient augmented Lagrangian based algorithm is developed to solve the above combined TV and nonlocal regularization constrained problem. Experimental results demonstrate that the proposed algorithm achieves significant performance improvements over the state-of-the-art TV based algorithm in both PSNR and visual perception.

*Index Terms*—Compressive sensing, augmented Lagrangian, total variation, image recovery, nonlocal regularization

## I. INTRODUCTION

The recent development of Compressive Sensing (CS) theory has drawn quite an amount of attention as an alternative to the current methodology of sampling followed by compression [1–5]. By exploiting the redundancy existed in a signal, CS conducts sampling and compression at the same time. From many fewer acquired measurements than suggested by the Nyquist sampling theory, CS theory demonstrates that, a signal can be reconstructed with high probability when it exhibits sparsity in some domain.

CS-based compression has an attractive advantage that the encoder is made signal-independent and computationally inexpensive at the cost of high decoder complexity, which is severely desirable in some image processing applications when the data acquisition devices must be simple (e.g. inexpensive resource-deprived sensors), or when oversampling can harm the object being captured (e.g. X-ray imaging) [2].

More specifically, a signal $u$ of size N is said to be sparse in domain $\Psi$, if its transform coefficients $\Psi u$ are mostly zeros, or nearly sparse if the dominant portion of coefficients are either zeros or very close to zeros. Given M linear measurements of $u$, denoted by $b$, the CS recovery of $u$ from $b$ is formulated as the following constrained optimization problem:

$$\min_{u} \|\Psi u\|_p \quad \text{s.t.} \quad b = Au, \qquad (1)$$

where $A$ represents random projections, $p$ is usually set to 1 or 0, characterizing the sparsity of the vector $\Psi u$. $\|*\|_1$ is $\ell_1$ norm, adding all absolute values of the entries in a vector, while $\|*\|_0$ is $\ell_0$ norm, counting the nonzero entries of a vector.

For image CS recovery, the above $u$ represents a column stacked image. The most current CS recovery algorithms explore the prior knowledge that a natural image is sparse in the some domains, such as DCT [4], wavelets [9], and gradient domain utilized by total variation (TV) model [5–8]. Particularly, TV based CS recovery methods achieve state-of-the-art results [8]. The TV based CS recovery model is defined as:

$$\min_{u} \|\mathcal{D}u\|_1 \quad \text{s.t.} \quad b = Au, \qquad (2)$$

where $\mathcal{D} = [\mathcal{D}_v, \mathcal{D}_h]$, and $\mathcal{D}_v, \mathcal{D}_h$ denote vertical and horizontal finite difference operators, respectively. The subject of solving (2) has received considerable interests, and many solvers are released publicly available (such as $\ell_1$-Magic [5], TwIST [6], NESTA [7], TVAL3 [8] and so on). Among them, TVAL3 [8] is faster and more effective than the other TV solvers, achieving the best CS recovery results.

Despite high effectiveness in image CS recovery, TV based algorithms often suffer from undesirable staircase artifact and tend to over-smooth image details and textures. To overcome this drawback, a new total variation measure TVL1 is designed, which enforces the sparsity and the directional continuity in the image gradient domain [14]. Lately, intra prediction is introduced to generate a sparser residual gradient domain, leading to an improved TV minimization algorithm for block-based image CS recovery [16]. Nonetheless, the improvements in these works are not satisfying, since only image local statistics are utilized neglecting image nonlocal statistics. Hence, many latest works concentrate on utilization of both local smoothness and nonlocal self-similarity for high quality image restoration and CS recovery [13, 15, 17].

In recent years, nonlocal means (NLM) filtering [10] has proven to be a useful tool in various image restoration tasks, such as image denoising, deblurring and super-resolution [11, 12]. By weighted filtering according to the degree of similarity among surrounding pixels, NLM filtering is very effective in sharpening edges and preserving image details, which essentially exploits the nonlocal self-similarity of natural images.

In this paper, an improved TV based image CS recovery algorithm is proposed by introducing a new NLM based nonlocal regularization constraint. To the best of our knowledge, this is the first time that NLM based nonlocal regularization is exploited in image CS recovery. Moreover, an efficient augmented Lagrangian based algorithm is developed to solve the above combined TV and nonlocal regularization constrained CS problem. The local TV model and nonlocal regularization are complementary to each other, making the proposed algorithm highly effective in preserving sharp image edges and more details while eliminating the staircase artifacts. Experimental results validate the performance of the proposed algorithm in both PSNR and visual perception.

The remainder of the paper is organized as follows. Section II gives the definition of our proposed NLM based nonlocal regularization. Section III describes the improved TV based image CS recovery algorithm by nonlocal regularization (hereinafter referred to as TVNLR) and provides the implementation details of TVNLR. Experimental results are reported in Section IV. In Section V, we conclude this paper.

## II. NLM BASED NONLOCAL REGULARIZATION

The basic idea of nonlocal means (NLM) filtering concept is very simple. For a given pixel $u_i$ in an image $\boldsymbol{u}$, its NLM filtered new intensity value, denoted by $\mathcal{NLF}(u_i)$, is obtained as a weighted average of surrounding pixels within a search window. The weight of $u_j$ to $u_i$, denoted by $\mathrm{w}_{ij}$, is determined by the similarity between these pixels, which is calculated by the Gaussian $\ell_2$-distance between the blocks centered at these pixels. Specifically, let $\boldsymbol{u}_i, \boldsymbol{u}_j$ denote the central pixel of $b_s \times b_s$ blocks $\boldsymbol{U}_i, \boldsymbol{U}_j$, and assume that $u_j$ lies in the $L \times L$ search window of $u_i$. Then, the weight $\mathrm{w}_{ij}$ is computed by

$$\mathrm{w}_{ij} = \exp(-\|\boldsymbol{U}_j - \boldsymbol{U}_i\|_2^2 / h^2) / c_i, \quad (3)$$

where $h$ is a controlling factor for Gaussian kernel and $c_i$ is the normalizing constant. According to the notations above, we generalize NLM filtering scheme to establish a novel nonlocal regularization (NR) for general CS recovery problems as

$$\mathcal{NR}(\boldsymbol{u}) = \sum_{u_i \in \boldsymbol{u}} \|u_i - \mathcal{NLF}(u_i)\|_2^2 = \sum_{u_i \in \boldsymbol{u}} \|u_i - \mathbf{w}_i^T \boldsymbol{\kappa}_i\|_2^2, \quad (4)$$

where $\boldsymbol{\kappa}_i$ is the column vector containing all the central pixels around $u_i$ in the search window and $\mathbf{w}_i$ is the column vector containing all the corresponding weights $\mathrm{w}_{ij}$. Furthermore, Eq. (4) can be rewritten in a matrix form as

$$\mathcal{NR}(\boldsymbol{u}) = \|\boldsymbol{u} - \boldsymbol{W}\boldsymbol{u}\|_2^2, \quad \boldsymbol{W}(i,j) = \begin{cases} \mathrm{w}_{ij}, & \text{if } u_j \in \boldsymbol{\kappa}_i \\ 0, & \text{otherwise} \end{cases}. \quad (5)$$

Note that the major difference between Eq. (5) and the proposed regularization for image deblurring by [12] is that the weight $\boldsymbol{W}$ in (5) is calculated by the original image while the weights in [12] is calculated by the observed blurred image. That means $\boldsymbol{W}$ are updated in (5) from iteration to iteration, rather than fixed in [12] at the beginning.

## III. TV BASED CS RECOVERY BY NONLOCAL REGULARIZATION

Incorporating (2) and (5) into CS optimization problem together, the proposed improved TV based image CS recovery algorithm by nonlocal regularization (TVNLR) is formulated as:

$$\min_{\boldsymbol{u}} \|\mathcal{D}\boldsymbol{u}\|_1 + \alpha \|\boldsymbol{u} - \boldsymbol{W}\boldsymbol{u}\|_2^2 \quad \text{s.t.} \quad \boldsymbol{A}\boldsymbol{u} = \boldsymbol{b}. \quad (6)$$

Note that Problem (6) is quite difficult to solve directly due to the non-differentiability and non-linearity of the combined regularization terms. Solving it efficiently is one of the main contributions of this paper. In this section, an augmented Lagrangian based approach is developed to solve Problem (6). The implementation details of TVNLR are given below.

At first, Problem (6) is transformed into an equivalent variant through variable splitting technique by introducing auxiliary variables $\boldsymbol{w}$ and $\boldsymbol{x}$:

$$\min_{\boldsymbol{w},\boldsymbol{u},\boldsymbol{x}} \|\boldsymbol{w}\|_1 + \alpha \|\boldsymbol{x} - \boldsymbol{W}\boldsymbol{x}\|_2^2 \quad (7)$$
$$\text{s.t.} \quad \mathcal{D}\boldsymbol{u} = \boldsymbol{w}, \boldsymbol{u} = \boldsymbol{x}, \boldsymbol{A}\boldsymbol{u} = \boldsymbol{b}.$$

Hence, the corresponding augmented Lagrangian function of Problem (7) is expressed as

$$\mathcal{L}_{\mathcal{A}}(\boldsymbol{w},\boldsymbol{u},\boldsymbol{x}) = \|\boldsymbol{w}\|_1 - \mathbf{v}^T(\mathcal{D}\boldsymbol{u} - \boldsymbol{w})$$
$$+ \tfrac{\beta}{2}\|\mathcal{D}\boldsymbol{u} - \boldsymbol{w}\|_2^2 + \alpha\|\boldsymbol{x} - \boldsymbol{W}\boldsymbol{x}\|_2^2 - \boldsymbol{\gamma}^T(\boldsymbol{u} - \boldsymbol{x}) \quad (8)$$
$$+ \tfrac{\theta}{2}\|\boldsymbol{u} - \boldsymbol{x}\|_2^2 + \tfrac{\mu}{2}\|\boldsymbol{A}\boldsymbol{u} - \boldsymbol{b}\|_2^2 - \boldsymbol{\lambda}^T(\boldsymbol{A}\boldsymbol{u} - \boldsymbol{b}),$$

where $\beta, \theta, \mu$ are regularization parameters associated with penalty terms $\|\mathcal{D}\boldsymbol{u} - \boldsymbol{w}\|_2^2, \|\boldsymbol{u} - \boldsymbol{x}\|_2^2, \|\boldsymbol{A}\boldsymbol{u} - \boldsymbol{b}\|_2^2$, respectively.

The basic idea behind the augmented Lagrangian method is to seek a saddle point of $\mathcal{L}_{\mathcal{A}}(\boldsymbol{w},\boldsymbol{u},\boldsymbol{x})$, which is also the solution of Problem (7). We utilize the augmented Lagrangian method to solve constrained Problem (7) by iteratively solving Problems (9) and (10):

$$(\boldsymbol{w}^{(k+1)}, \boldsymbol{u}^{(k+1)}, \boldsymbol{x}^{(k+1)}) = \underset{\boldsymbol{w},\boldsymbol{u},\boldsymbol{x}}{\arg\min}\, \mathcal{L}_{\mathcal{A}}(\boldsymbol{w},\boldsymbol{u},\boldsymbol{x}), \quad (9)$$

$$\begin{cases} \mathbf{v}^{(k+1)} = \mathbf{v}^{(k)} - \beta(\mathcal{D}\boldsymbol{u}^{(k+1)} - \boldsymbol{w}^{(k+1)}), \\ \boldsymbol{\gamma}^{(k+1)} = \boldsymbol{\gamma}^{(k)} - \theta(\boldsymbol{u}^{(k+1)} - \boldsymbol{x}^{(k+1)}), \\ \boldsymbol{\lambda}^{(k+1)} = \boldsymbol{\lambda}^{(k)} - \mu(\boldsymbol{A}\boldsymbol{u}^{(k+1)} - \boldsymbol{b}). \end{cases} \quad (10)$$

Here, the subscript $k$ denotes the iteration number, and $\mathbf{v}, \boldsymbol{\gamma}, \boldsymbol{\lambda}$ are the Lagrangian multipliers associated with the constraints $\mathcal{D}\boldsymbol{u} = \boldsymbol{w}, \boldsymbol{u} = \boldsymbol{x}, \boldsymbol{A}\boldsymbol{u} = \boldsymbol{b}$, separately.

It is evident that Problem (9) is still hard to solve efficiently in a direct way owing to its non-differentiability. To avoid this problem, a quite useful alternating direction technique [8] is employed, which alternatively minimizes one variable while fixing the other variables. Due to the alternating direction technique, Problem (9) is decomposed into the ensuing three sub-problems. In the following, we argue that the every separated sub-problem admits an efficient solution. For simplicity, the subscript $k$ is omitted without confusion.

### A. $\boldsymbol{w}$ Sub-problem

Given $\boldsymbol{u}, \boldsymbol{x}$, after simplifications, the optimization problem associated with $\boldsymbol{w}$ can be expressed as

$$\min_{\boldsymbol{w}} \tfrac{\beta}{2}\|\mathcal{D}\boldsymbol{u} - \boldsymbol{w}\|_2^2 - \mathbf{v}^T(\mathcal{D}\boldsymbol{u} - \boldsymbol{w}) + \|\boldsymbol{w}\|_1. \quad (11)$$

According to the lemma in [8], the closed form of (11) is

$$\tilde{\boldsymbol{w}} = \max\left\{\left|\mathcal{D}\boldsymbol{u} - \tfrac{\mathbf{v}}{\beta}\right| - \tfrac{1}{\beta}, 0\right\} \cdot \mathrm{sgn}\left(\mathcal{D}\boldsymbol{u} - \tfrac{\mathbf{v}}{\beta}\right). \quad (12)$$

## B. $u$ Sub-problem

With the aid of $w, x$, the $u$ sub-problem is equivalent to

$$\min_{u} \begin{cases} -\mathbf{v}^T(\mathcal{D}u-w)+\frac{\beta}{2}\|\mathcal{D}u-w\|_2^2-\boldsymbol{\gamma}^T(u-x) \\ +\frac{\theta}{2}\|u-x\|_2^2-\boldsymbol{\lambda}^T(Au-b)+\frac{\mu}{2}\|Au-b\|_2^2 \end{cases}. \quad (13)$$

Clearly, Eq. (13) is a quadratic function and its gradient $d$ is simplified as $d = \mathcal{D}^T(\beta\mathcal{D}u-\mathbf{v}-\beta w)-\boldsymbol{\gamma}+\theta(u-x)+A^T(\mu(Au-b)-\boldsymbol{\lambda})$. Thus, setting $d=0$ can give us the exact minimizer of Problem (13), that is,

$$u = G^{-1}\left(\mathcal{D}^T\mathbf{v}+\boldsymbol{\gamma}+\theta x+A^T\boldsymbol{\lambda}+\mu A^T b+\beta\mathcal{D}^T w\right), \quad (14)$$

where $G = (\beta\mathcal{D}^T\mathcal{D}+\theta I+\mu A^T A)$ and $I$ is identity matrix. However, computing the inverse of $G$ at each iteration is too costly to implement numerically. Therefore, an iterative method is highly desirable. Here, the steepest descent method with the optimal step is used to solve (13) iteratively by applying

$$\tilde{u} = u - \eta d, \quad (15)$$

where $\eta = \mathrm{abs}(d^T d / d^T G d)$ is the optimal step, Thus, solving $u$ requires computing (15) from iteration to iteration.

## C. $x$ Sub-problem

Given $w, u$, similarly, the $x$ sub-problem becomes

$$\min_{x} \frac{\theta}{2}\|u-x\|_2^2 - \boldsymbol{\gamma}^T(u-x) + \alpha\|x-Wx\|_2^2. \quad (16)$$

Problem (16) can be further simplified into

$$\min_{x} \frac{1}{2}\|x-r\|_2^2 + \frac{\alpha}{\theta}\|x-Wx\|_2^2, \quad (17)$$

where $r = (u - \frac{\gamma}{\theta})$. Next, the optimal weight $W$ is updated using the last estimate of $x$. Then, the exact solution of (17) can be obtained by setting the gradient of (17) to zero, i.e.,

$$x = H^{-1}r, \quad (18)$$

where $H = I + \frac{2\alpha}{\theta}(I-W)^T(I-W)$. Nonetheless, analogous to (14), calculating the inverse of $H$ is too costly. In order to overcome this disadvantage of high complexity, we simplify (17) and design a novel and efficient solution, rather than adopt an iterative approach.

Concretely, we start from the basic fact that the already known $r$ in (17) can be viewed as a approximate observation of $x$. That means both $x$ and $r$ have the same low frequency components (LFC). Since the weight matrix $W$ actually represents the nonlocal means denoising operator, we approximately have the equation $Wx = Wr$, which stands for the denoised versions of $x, r$, corresponding to LFC of $x, r$, respectively. Accordingly, Problem (17) can be written as

$$\min_{x} \frac{1}{2}\|x-r\|_2^2 + \frac{\alpha}{\theta}\|x-Wr\|_2^2. \quad (19)$$

Finally, setting the gradient of (19) to zero, we acquire the closed-form solution of $x$ sub-problem (17) as follows

$$\tilde{x} = \frac{\theta r + 2\alpha Wr}{\theta + 2\alpha}. \quad (20)$$

So far, all issues for handing the sub-problems have been solved. In fact, we achieve efficient solutions for each separated sub-problem, which enables the whole algorithm more efficient and effective. In light of all derivations above, the complete description of TVNLR is stated clearly in Table I.

TABLE I The Complete Description of Proposed TVNLR

**Input:** The observed measurement $b$, the measurement matrix $A$ and $\beta, \mu, \theta, \alpha$.
Initialization: $u_0 = A^T b$, $\mathbf{v}_0 = \boldsymbol{\gamma}_0 = \boldsymbol{\lambda}_0 = \mathbf{0}$; $w_0 = x_0 = \mathbf{0}$.
 **while** Outer stopping criteria unsatisfied **do**
  **while** Inner stopping criteria unsatisfied **do**
   Solve $w$ sub-problem by computing Eq. (12);
   Solve $u$ sub-problem by computing Eq. (15);
   Compute the weight $W$ using Eq. (3);
   Solve $x$ sub-problem by computing Eq. (20);
  **end while**
  Update multipliers $\mathbf{v}, \boldsymbol{\gamma}, \boldsymbol{\lambda}$ by computing Eq. (10);
 **end while**
**Output:** Final CS restored image $\tilde{u}$.

## IV. EXPERIMENTAL RESULTS

In this section, experimental results on four conventional natural images are presented to evaluate the performance of the proposed TVNLR. In our experiments, the CS measurements are obtained by applying a Gaussian random projection matrix to the original image signal, where $A$ in Eq. (6) is generated by Matlab command **randn**(M,N) (*ratio*=M/N). We empirically set $\mu = 128$, $\theta = 2$, $\beta = 32$, $\alpha = 16$, $b_s = 7$, $L = 13$ and $h = 0.03$ for all test images. As we know, finding suitable values for regularization parameters is crucial for image inverse problems. It is necessary to stress that the choice for all the parameters in this paper is general, and can be generalized to other natural images, which has been verified in our experiments. For other comparative methods, we choose the empirical parameter settings, which give the best performance in the recovery quality. The experimental environment is a Dell OPTIPLEX computer with Intel(R) Core(TM) 2 Duo CPU (3.00GHz) and 4G memory, running Windows XP and Matlab R2011a.

The proposed TVNLR is compared with two representative CS recovery methods in literature, i.e., tree-structured DCT (TSDCT) method [4] and total variation (TVAL3) method [8], which deal with the image signal in the DCT domain and the gradient domain, respectively. It is worth emphasizing that TVAL3 method is known as one of the state-of-the-art algorithms for image CS recovery.

TABLE II PSNR Comparison of Various Algorithms (Unit: dB)

| *Image* | *Barbara* (256×256) | | | | *Leaves* (256×256) | | | |
|---|---|---|---|---|---|---|---|---|
| ratio (%) | 15 | 20 | 25 | 30 | 15 | 20 | 25 | 30 |
| **TSDCT[4]** | 21.25 | 22.11 | 23.35 | 25.01 | 17.66 | 19.26 | 20.79 | 22.29 |
| **TVAL3[8]** | 23.46 | 24.33 | 25.32 | 26.32 | 21.18 | 23.11 | 24.80 | 26.53 |
| **TVNLR** | 24.25 | 25.66 | 26.89 | 28.39 | 22.27 | 24.44 | 26.21 | 28.41 |
| **Gain** | **0.79** | **1.33** | **1.57** | **2.07** | **1.09** | **1.33** | **1.41** | **1.88** |
| *Image* | *Lena* (256×256) | | | | *House* (256×256) | | | |
| ratio (%) | 15 | 20 | 25 | 30 | 15 | 20 | 25 | 30 |
| **TSDCT[4]** | 24.43 | 25.35 | 26.53 | 27.96 | 28.86 | 30.59 | 31.46 | 32.47 |
| **TVAL3[8]** | 27.47 | 28.85 | 30.06 | 31.26 | 31.74 | 33.07 | 34.15 | 35.02 |
| **TVNLR** | 28.39 | 29.84 | 31.17 | 32.33 | 33.24 | 34.24 | 35.38 | 36.13 |
| **Gain** | **0.92** | **0.99** | **1.11** | **1.07** | **1.50** | **1.17** | **1.23** | **1.11** |

The PSNR results reconstructed by three comparative methods under different *ratios* for all test images are provided in Table II. The second row of Table II denotes the CS measurements *ratio* and the last row denotes the PSNR gain achieved by proposed TVNLR over the traditional TV based CS recovery method TVAL3. It is clear to see that TSDCT [4] obtains the lowest PSNR. TVAL3 acquires much better PSNR results than TSDCT. The proposed TVNLR considerably outperforms the other methods in all the cases, with PSNR improvements of up to 2.07 dB and 5.4 dB, compared with TVAL3 and TSDCT, respectively. Furthermore, the average PSNR gain earned by TVNLR over TVAL3 is 1.2 dB, which fully demonstrates the superiority of the proposed TVNLR over the traditional TV based CS recovery algorithms.

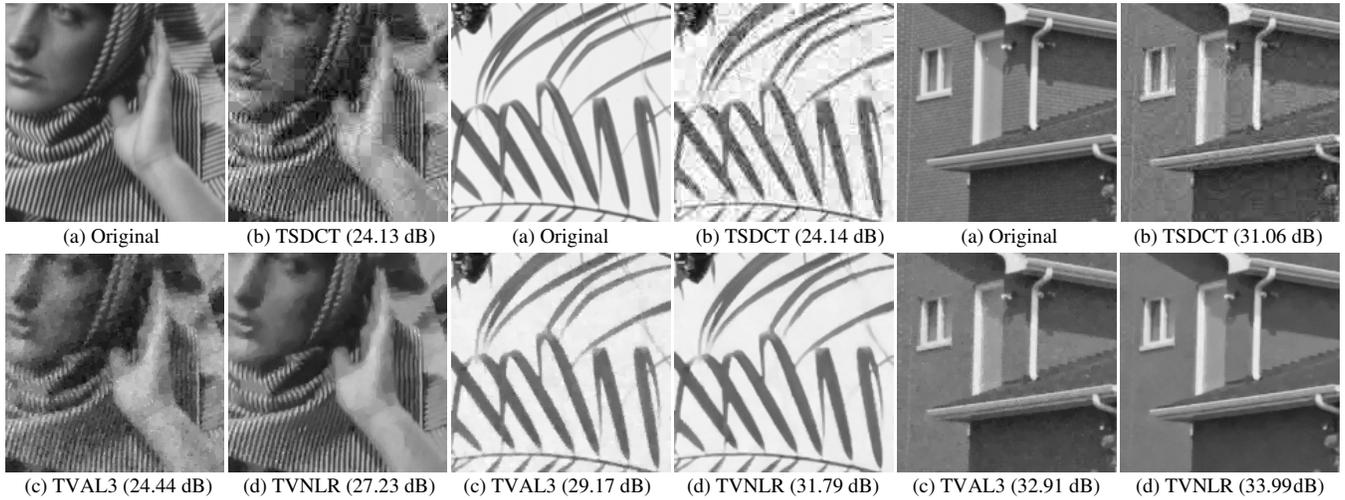

**Fig. 1.** Visual quality comparison of portions of CS recovered Images *Barbara*, *Leaves* and *House*. (The measurements *ratio* is *30%*). (a) Original images; (b)–(d) Image CS recovered results by TSDCT [4], TVAL3 [8], and the proposed TVNLR. **Please enlarge and view the figures on the screen for better comparison.**

Some visual results of portions of CS recovered images for the three comparative algorithms in the case *ratio=30%* are presented in Fig. 1. Obviously, TSDCT generates the worst perceptual results (see Fig. 1(b)). The CS recovered images by TVAL3 possess much better visual quality than TSDCT, but still suffer from some undesirable artifacts, such as edge over-smoothness and staircase effects (see Fig. 1(c)). The proposed algorithm TVNLR not only eliminates the staircase effects (i.e., wall in Image *House*) and preserves sharp edges (i.e., leaf in Image *Leaves*) but also recovers more detail of textures in images (i.e., scarf in Image *Barbara*), exhibiting the best visual quality (see Fig. 1(d)).

Table III provides the average computational time of various algorithm for all the test images in each ratio. With the inclusion of nonlocal regularization, additional computational overhead is added to TVNLR. On average, the time computational complexity of TVNLR is rough 2.4 times that of TVAL3.

TABLE III  Average Computational Time for Various Algorithms (Unit: s)

| ratio (%) | 15 | 20 | 25 | 30 | Avg. |
|---|---|---|---|---|---|
| **TSDCT [4]** | 1308 | 1513 | 1706 | 1611 | **1612** |
| **TVAL3 [8]** | 330 | 389 | 432 | 481 | **408** |
| **TVNLR** | 839 | 971 | 998 | 967 | **944** |

## V. Conclusion

This paper presents an improved TV based image CS recovery algorithm by introducing a new NLM based nonlocal regularization constraint. The local TV model and nonlocal regularization are complementary to each other, making the proposed algorithm highly effective in preserving sharp image edges and more details while eliminating the staircase artifacts. Experimental results validate the performance of the proposed algorithm in both PSNR and visual perception.

## Acknowledgment

We would like to thank the authors of [4] and [8] for kindly providing their codes.